\documentclass[11pt]{article}

\usepackage[final]{acl}

\usepackage{times}
\usepackage{latexsym}

\usepackage[T2A,T1]{fontenc}
\usepackage[utf8]{inputenc}
\usepackage[russian,english]{babel}

\usepackage{microtype}

\usepackage{inconsolata}

\usepackage{gb4e}
\noautomath % Prevents gb4e from interfering with math symbols

\usepackage{booktabs}
\usepackage{multicol}
\usepackage{multirow}
\usepackage{graphicx}

\let\svthefootnote\thefootnote
\newcommand\freefootnote[1]{%
  \let\thefootnote\relax%
  \footnotetext{#1}%
  \let\thefootnote\svthefootnote%
}

\usepackage{color-edits}
\addauthor{gn}{magenta}
\addauthor{lt}{cyan}
\addauthor{ll}{orange}
\addauthor{ap}{blue}
\addauthor{mg}{red}

\title{GlossLM: A Massively Multilingual Corpus and Pretrained Model for Interlinear Glossed Text}

\author{Michael Ginn*${ }^{1}$ \quad Lindia Tjuatja*${ }^{2}$ \quad  Taiqi He${ }^{2}$  \quad
       Enora Rice${ }^{1}$ \\ {\bf Graham Neubig}${ }^{2}$ \quad {\bf Alexis Palmer}${ }^{1}$ \quad {\bf Lori Levin}${ }^{2}$\\
    ${ }^{1}$University of Colorado Boulder \quad ${ }^{2}$Carnegie Mellon University  \\ \texttt{michael.ginn@colorado.edu}  \quad \texttt{lindiat@andrew.cmu.edu} \\ $*$ Equal contribution}

\begin{document}
\maketitle
\begin{abstract}
Language documentation projects often involve the creation of annotated text in a format such as \textbf{interlinear glossed text (IGT)}, which captures fine-grained morphosyntactic analyses in a morpheme-by-morpheme format. However, there are few existing resources providing large amounts of standardized, easily accessible IGT data, limiting their applicability to linguistic research, and making it difficult to use such data in NLP modeling. 

We compile the largest existing corpus of IGT data from a variety of sources, covering over 450k examples across 1.8k languages, to enable research on crosslingual transfer and IGT generation. We normalize much of our data to follow a standard set of labels across languages.

Furthermore, we explore the task of automatically generating IGT in order to aid documentation projects. As many languages lack sufficient monolingual data, we pretrain a large multilingual model on our corpus. We demonstrate the utility of this model by finetuning it on monolingual corpora, outperforming SOTA models by up to 6.6\%. Our pretrained model and dataset are available on Hugging Face.\footnote{\url{https://huggingface.co/collections/lecslab/glosslm-66da150854209e910113dd87}}

%as well as provide access through a web interface for use in language documentation efforts.

% Prior work has explored methods to automatically generate IGT in order to reduce the time cost of language analysis. However, many languages (particularly those requiring preservation) lack sufficient IGT data to train effective models, and crosslingual transfer has been proposed as a method to overcome this limitation.

% Then, we pretrain 

\end{abstract}

\section{Introduction}

% \gncomment{In section titles, title case and initial upper case only are used inconsistently. I would choose one and stick with it (probably title case, unless the section title is a complete sentence).}

% \ltcomment{TODO: main figure---color-coded IGT example, show that our system generates the gloss line}
With nearly half of the world's 7,000 languages considered endangered, communities of minoritized language speakers are working to preserve and revitalize their languages \citep{seifart_language_2018}. These efforts often involve collection, analysis, and annotation of linguistic data. Annotated text can be used in the creation of reference materials (such as dictionaries and grammars) as well as to develop language technologies including searchable digital text~\cite{blokland2019using, rijhwani-etal-2023-user} and computer-assisted educational tools~\cite{uibo_building_2017,chaudhary-etal-2023-teacher}. 
% \gncomment{Not a rush, but eventually let's try to add a few citations that are not our papers as well.}. 
%\apcomment{agreed, and it would be nice to include some work using IGT for such resources that is not coming from the NLP perspective; look through proceedings of ComputEL and ICLDC. I'll also see what I have to add.}

\begin{figure}[t]
  \centering
   \includegraphics[width=\linewidth]{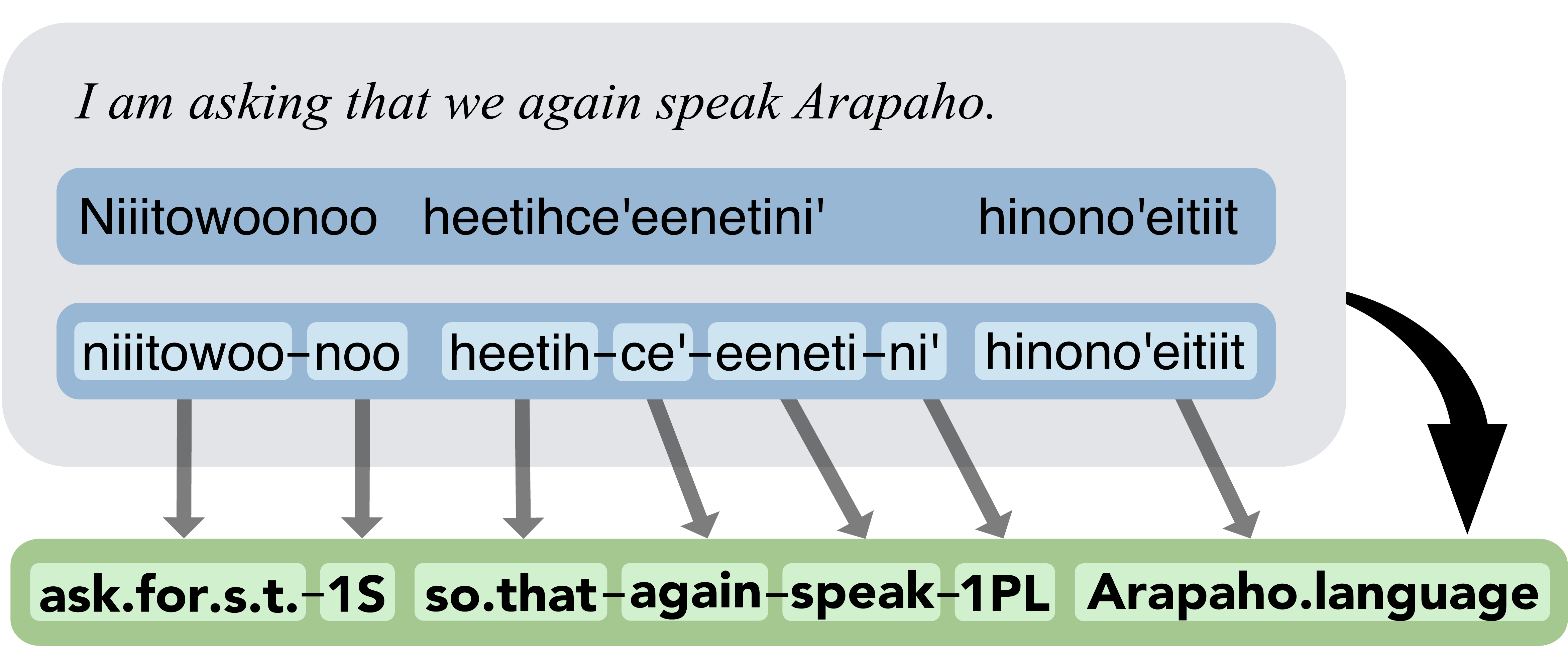}
  \caption{Components of interlinear gloss with an Arapaho sentence and English translation~\cite{cowell2020}. Blue boxes show transcriptions that are \textit{unsegmented} (top) or \textit{segmented} (bottom). Segmented text is split into morphemes which are aligned with the gloss labels shown in the green box. The task of automatic glossing uses some or all of the information in the gray box (transcription \& translation) to generate the gloss line.} 
  \label{fig:gloss-example}
\end{figure}

Interlinear glossed text (IGT) is a widespread format in language documentation for linguistic annotation. IGT is a multi-line data format (see \autoref{fig:gloss-example}) which includes (1) a transcription of speech in the language, (2) an aligned morpheme-by-morpheme description, and oftentimes (3) a free translation. IGT can be used to illustrate morphosyntactic features of languages that other researchers may not be familiar with, and it is a popular format for examples in linguistics papers and textbooks. It also serves as a resource in the NLP context for the creation of morphological paradigms \citep{moeller-etal-2020-igt2p}, machine translation \citep{Zhou2019UsingIG}, generating precision grammars \citep{bender-etal-2013-towards}, and other tools including POS taggers and dependency parsers \citep{georgi2016aari}.
%for the language, among other applications.
% \gncomment{Maybe add one more sentence connecting this back, such as ``These glossed texts can, in turn, be used as examples for grammatical sketches or language learning textbooks for the language.''}

% Linguistic fieldwork and documentation play a critical role in both building universal linguistic theories and preserving the world's linguistic diversity. A recent study by~\citet{bromham2022global} estimates that without intervention, the rate of language death could triple over the next 40 years---amounting to over one language lost every month---indicating an urgent need to document and record these languages. Despite best efforts in data collection in these languages, which usually takes the form of speech recordings and transcriptions, there is still a large bottleneck in data annotation and analysis due to the time-intensive nature and linguistic expertise required of such tasks ~\cite{seifart_language_2018}. One way to overcome this bottleneck is to create computational tools to assist linguists in annotation~\cite{palmer2009evaluating}. 

\paragraph{Compiling IGT Data} Though IGT often follows a common glossing format, gloss conventions vary wildly. Furthermore, IGT data is rarely compiled into large, standardized corpora, often existing as scattered examples in research papers. To address this, we compile the largest corpus of digitized IGT from various existing sources, with over 450k examples in 1.8k languages (\S\ref{sec:corpus}). We explore methods to normalize this data (\S\ref{sec:normalizing})., standardizing over 80\% of the grammatical glosses in the corpus to follow the UniMorph schema \citep{sylak2016composition}. We are releasing our corpora for future NLP, linguistics, and language documentation research.

\paragraph{Automating IGT Generation} The creation of new IGT corpora is often difficult and time-consuming, requiring documenters to perform linguistic analysis and extensive documentation simultaneously.
Research has found that computational tools can help accelerate annotation and overcome this bottleneck~\cite{palmer2009evaluating} by predicting the gloss line of IGT given a transcription.

The majority of prior work on automatic glossing focuses on training monolingual models that can predict IGT for a single language \citep{moeller_automatic_2018, mcmillan-major_automating_2020, zhao_automatic_2020}, however, these models can struggle when data is limited and require dedicated effort to train and deploy.
% Furthermore, language- and dataset-specific models are not easily adaptable to new data. 

% \gncomment{We don't have a paragraph on the status quo on IGT generation. We should cite previous work and point out the problems with it.}
We aim to overcome the monolingual data bottleneck by creating a \textbf{multilingual pretrained glossing model} that can be adapted to specific languages and gloss formats. We continually pretrain a model on our corpus, and find that the pretrained multilingual model retains high performance across languages. We then finetune the pretrained model on monolingual data, including languages that do not appear in the pretraining corpus. Our models achieve new SOTA performance on five out of seven languages, and demonstrate clear improvements for low-resource language settings over an equivalent finetuned model without our continual pretraining (\S\ref{sec:glosslm-experiments}).

\section{Interlinear Glossed Text (IGT)}
% \gncomment{I asked GPT-4 to create a first draft of this section and lightly modified, feel free to change.}
% Interlinear glossed text (IGT) is a crucial tool in linguistic research, especially for illustrating the grammatical structure of less widely known languages. It typically involves presenting a line of original text, followed by a word-by-word breakdown and a free translation. This method helps in understanding the morphological and syntactic properties of languages. 

% Here is an example of how IGT is used to present linguistic data:

% \begin{exe}
%   \ex
%   \gll Der Mann gibt dem Kind das Buch.\\
%        the.NOM man.NOM gives the.DAT child.DAT the.ACC book.ACC.\\
%   \glt `The man gives the child the book.'
% \end{exe}

% Each token in the original sentence (e.g.~Der) is associated with a word in the meta-language (e.g.~the) and grammatical tags (e.g.~NOM).
% This is often also associated with a free-text translation.
\subsection{Format}
Interlinear glossed text is a structured data format which presents text in a language being studied along with \textit{morphological glosses}---aligned labels that indicate each morpheme's meaning and/or grammatical function. Often, a free translation in a widely-spoken language is included as well. An IGT example for Arapaho is given in \autoref{ex:arapaho-gloss}~\cite{cowell2020}, with glosses and translations in English.

% \newpage %to keep example in one col
\begin{small}
\begin{exe}
  \ex
  \gll nuhu' tih-'eeneti-3i' heneenei3oobei-3i' \\
       this when.PAST-speak-3PL IC.tell.the.truth-3PL \\
  \glt ``When they speak, they tell the truth.''
  \label{ex:arapaho-gloss}
\end{exe}
\end{small}

\noindent This example is \textit{segmented}, with morphemes separated by dashes. Each morpheme in the Arapaho sentence (e.g.\textit{~tih}) is directly aligned with a gloss (e.g.~when.\textsc{Past}) that describes the morpheme's function and/or meaning. Labels in all caps (e.g.~\textsc{Past}) are grammatical glosses; lowercase labels (e.g.~speak) are lexical glosses. Periods are used for \textit{fusional morphemes}, which carry several morphological or lexical functions in a single morpheme.
% \apmargincomment{four examples are given in this paragraph, two in italics, two not - I'd say pick one and stick with it} \ltmargincomment{for now, using italics for transcriptions and non-italics for labels}

IGT examples may instead use \textit{unsegmented} transcriptions, as in the Uspanteko example in \autoref{ex:uspanteko-gloss}~\cite{okma}. 

\begin{small}
\begin{exe}
  \ex
  \label{ex:uspanteko-gloss}
  \gll o sey xtok rixoqiil \\
       o sea COM-buscar E3S-esposa \\
  \glt ``O sea busca esposa.''
\end{exe}
\end{small}

\noindent Here, words and their labels are aligned, but no explicit alignment between morpheme glosses and individual morphemes is provided, and thus the segmentation of words into morphemes is unclear.

\subsection{Challenges with Interlinear Glossing}
% \gncomment{I usually try to avoid having a single sub-section in a section. You could possibly make the first part of this section a subsection as well.}

Effective glossing requires expert knowledge of the target language and linguistic understanding of morphological patterns. Furthermore, certain factors exist that make this task particularly difficult for automated systems. Often, transcriptions are not segmented into morphemes, and systems must perform simultaneous segmentation and glossing.

Glossing conventions and formats may vary widely from documenter to documenter \cite{chelliah_using_2021}, with differences in label spelling (e.g. \textsc{Sing}/\textsc{Sg}/\textsc{S} to denote singular), formatting and punctuation, and language-specific labels. Furthermore, nearly all languages have very little digitized IGT data, posing difficulty to automated systems and human annotators alike. Finally, even when automated systems have been created, practical deployment remains an additional challenge for documenters.

\section{GlossLM Corpus}
\label{sec:corpus}
While various publicly available sources of digitized IGT exist, the lack of unified data formatting and ease of access is a roadblock to using these resources effectively. To solve this problem, we compile and clean the largest IGT dataset from a variety of sources and languages. In total, our dataset contains over 450k IGT instances (from 250k unique sentences) in 1.8k languages, collected from six different IGT corpora. All sources are publicly available under the CC BY 4.0 License, allowing free use and redistribution, and we have confirmed with the creators of each source that our usage is within the intended use. We make our artifacts available under the same license.
\subsection{Data Sources}
\label{sec:datasources}
\begin{table}[h]
\small
% \begin{minipage}{0.9\columnwidth}
    \centering
    \begin{tabular}{lcc}
    \toprule
    Corpus & Languages & IGT instances \\
    \midrule
    ODIN & 936 & 83,661 \\
    SIGMORPHON & 7 & 68,604  \\
    IMTVault & 1,116 & 79,522  \\
    APICS & 76 & 15,805 \\
    UraTyp & 35 & 1,719 \\
    Guarani Corpus & 1 & 803 \\
    \midrule
    Total & 1,782 & 250,582 \\
    \bottomrule
    \end{tabular}%
    \caption{Number of unique examples and languages in each source corpus for the \textsc{GlossLM} dataset.} 
    % If an IGT example provides segmented text, we also create an unsegmented example.
    % \end{minipage}
    \label{tab:dataset}
\end{table}

\paragraph{ODIN}
The Online Dictionary of Interlinear Text (ODIN, \citealt{lewis_developing_2010}) is a large dataset of 158k IGT examples representing 1496 languages, compiled by scraping IGT from linguistics documents on the internet. We use the preprocessed version of ODIN by~\citet{he-etal-2023-sigmorefun}, which discards languages with fewer than five IGT samples, resulting in 84k unique glossed sentences across 936 languages.

\paragraph{SIGMORPHON Shared Task} We use the IGT data from the 2023 SIGMORPHON Shared Task on Interlinear Glossing \citep{ginn-etal-2023-findings}. The data covers seven languages with diverse features and includes 69k glossed sentences. We use the shared task corpora as our primary evaluation sets, with the same splits as the shared task.

\paragraph{IMTVault} IMTVault \citep{nordhoff-kramer-2022-imtvault} is a recent aggregation of IGT data extracted from \LaTeX{} code in books published by the Language Science Press. We use the 1.1 release \citep{sebastian_nordhoff_2023_10113016} which includes 1116 languages and 80k examples.

\paragraph{APiCS}
The Atlas of Pidgin and Creole Language Structures (APiCS) is a set of books detailing grammatical features of 76 pidgin and creole languages~\citep{Michaelis2013Atlas, Michaelis2013Survey}.  APiCS online provides interactive versions of the books, including 16k IGT examples.

\paragraph{UraTyp}
UraTyp~\citep{miina_norvik_2022_6392555} provides grammatical and typological information, collected from linguistic questionnaires on various languages. This includes a small number of IGT examples (1.7k) spanning 35 languages.

\paragraph{Guarani Corpus}
The Guarani Corpus~\citep{GuaraniCorpus} consists of 803 examples of IGT, representing fifteen stories, for Guarani, a Tupian language spoken in South America. We use Beautiful Soup\footnote{\url{https://pypi.org/project/beautifulsoup4/}} to parse examples from HTML.

\subsection{Preprocessing}
In total we have 250k unique IGT instances in 1.8k languages. 
% We use Hugging Face Datasets to store our corpus in a widely-used, publicly accessible format. 
% Each row retains a unique ID and reference to its source, so examples can be easily cross-referenced.
If datasets explicitly indicate whether an IGT line is segmented, we use this value. Otherwise, we determine segmentation by checking if a line has any morpheme boundary markers (the hyphen~``-''). For segmented words, we remove the segmentation markers to create an additional unsegmented version of the same example, for a total of 451k examples (206k segmented).

We run \texttt{langid} \citep{lui-baldwin-2012-langid} on translations to verify the translation language labels, and leave the translation field blank if the predicted language did not match the language indicated by the original source. Finally, we pad any non-lexical punctuation with spaces and normalize spacing, as our experiments indicate that our models are sensitive to this formatting.

\subsection{Language Coverage}

Within our dataset, around 90\% of examples have an associated Glottocode~\citep{hammarstrom2023glottolog}, amounting to 1,785 unique Glottocodes and over 150 language families represented. While it would be ideal to have a relatively balanced set across languages and language families, many languages only have a few lines of IGT available, and thus our dataset has a long tail distribution across languages. The language with the greatest representation by far is Arapaho (from the SIGMORPHON Shared Task dataset) with almost 98k IGT instances, making up about 20\% of the entire dataset. Overall, 25\% of languages have fewer than 5 IGT instances, 50\% have fewer than 10, and 75\% have fewer than 54. We include histograms for the distributions across languages and language families  in~\autoref{sec:langdist}, as well as preliminary analysis of typological coverage using the Grambank database~\cite{grambank_release} in~\autoref{sec:grambank}.

\section{Normalizing Gloss Labels}
\label{sec:normalizing}
\subsection{Motivation}
As our data comes from a variety of sources, spanning many languages and documentation projects, there is a great amount of diversity in the morphological glosses used. This includes cases where several different labels are used to indicate the same feature (e.g. \textsc{Sing}, \textsc{Sg}, or \textsc{S} for singular), as well as formatting differences such as the usage of periods (e.g. \textsc{1Sg} vs \textsc{1.Sg}).

We explore the feasibility and value of normalizing glosses to a single standardized format. On one hand, normalizing glosses may make it easier to train models that utilize crosslingual information through shared gloss labels, but on the other hand, it is difficult (perhaps impossible) to select a single schema that preserves the original intent of all annotators.

\begin{figure}[h]
    \centering
    \includegraphics[width=\linewidth]{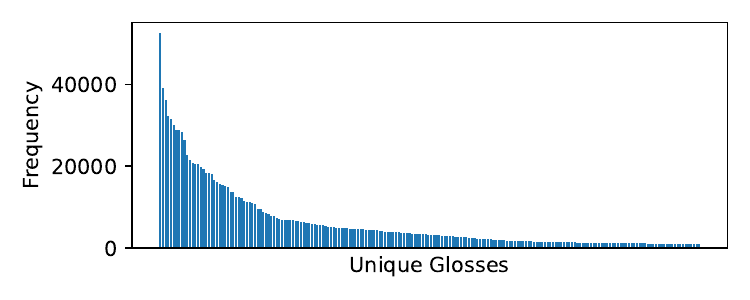}
    \caption{Distribution of unique glosses across all languages.}
    \label{fig:glosses}
\end{figure}

We split gloss lines by period and count the number and frequency of unique grammatical gloss labels across our corpus (focusing on the all-caps functional glosses, not stem translations) and visualize the distribution in \autoref{fig:glosses}.

There are 11,493 unique glosses which roughly form a Zipfian distribution \citep{zipf1945}. The most common glosses, unsurprisingly, are labels such as \textsc{Pl} (plural, 52,488 instances), \textsc{3Sg} (3rd-person singular, 39,147), and \textsc{Past} (36,124), which occur broadly across many languages. 

Normalizing all unique glosses would be a monumental task with uncertain benefits. However, we observe that the 200 most common gloss types account for 82.7\% of glosses in our dataset. We focus on normalizing these glosses: e.g. all instances of \textsc{Past} and \textsc{Pst}, which are both in the top 200, should be normalized to the same label. We note that there are other aspects of the data that vary (e.g. periods vs. underscores for multi-word glosses, representing non-concatenative morphology), as addressed in~\citet{mortensen-etal-2023-generalized}. Future work could potentially focus on the benefits of these aspects of normalization on training.

\subsection{Methodology}
We select the UniMorph schema of \citet{sylak2016composition} as our standardized set. While no single set of labels captures the intricacies of all of the world's languages, UniMorph is widely used and has coverage for many common features. 

The two lead authors of this paper jointly created a mapping from the labels in our dataset to UniMorph labels. While many mappings were obvious (or already compatible with UniMorph), others posed a myriad of issues. For glosses primarily used for a single language, we consulted the original source dataset to determine the meaning. 

\paragraph{Ambiguous labels} Several labels were ambiguous,  corresponding to one of several UniMorph glosses depending on the language and annotator. For example, the label \textsc{S} (appearing 20,855 times) is used for singular, subject, and noun/sustantivo (at least) in our dataset. In order to map these, we would need to analyze their meaning on a sentence-by-sentence basis, which was not feasible; thus, we left such ambiguous labels as-is.

\paragraph{Glosses not in UniMorph} UniMorph primarily focuses on common crosslingual inflectional features, and does not cover the full extent of the morphological systems of the world. We observed 64 of 200 (32\%) gloss labels with no clear UniMorph equivalent, including demonstratives (\textsc{Dem}, 15,585 instances), obliques (\textsc{Obl}, 13,639), and clitics (\textsc{Cl}, 7,453). In many cases, there is a related UniMorph gloss that is more general or more specific; for example preterite (\textsc{Pret}, 1,986) could be mapped to simple past (\textsc{Pst}). However, this would be an imprecise mapping, and could be confusing to a linguist of the particular language, so we again elect to leave these glosses unmapped.

We use our mapping to normalize the dataset and make the normalized version available in addition to the original.

\subsection{Use in Future Research}
We believe our dataset can potentially be useful across NLP research, linguistic research, and language documentation. NLP researchers benefit from a single, easily-accessible dataset covering many languages, which can be used for future research on interlinear glossing, morpheme analysis, segmentation, and translation. 

Linguistics researchers will be able to use the dataset to easily search for phenomena across languages, particularly with the normalized version of the dataset. For example, a linguist could find examples of sentences demonstrating the ergative/absolutive distinction. They could further refine this analysis by narrowing the results to a set of related languages, using the glottocodes in our dataset.

As another example, if a linguist wished to determine how prior researchers have annotated examples in a particular language, they would previously have to search across research papers, textbooks, and small corpora. With our dataset, it is trivial to pull up all of the examples in a particular language, potentially compiled from many sources.
% \footnote{\url{https://huggingface.co/datasets/lecslab/glosslm-corpus-split-unimorph} \apcomment{anonymize link }}

% \subsection{Variance in Glossing Conventions}
% \ltcomment{Maybe some discussion here with pointers to the appendix for more details. If we end up doing some normalization, we can merge this with the preprocessing subsection}

% \ltedit{As previously noted, one difficulty with automatic interlinear glossing is the inherent variance in glossing styles across sources and linguists. While the general formatting of IGT is largely consistent, there may be differences in the form of labels (e.g. SING/SG/S) and other low-level formatting differences like the usage of periods (e.g. 1SG vs. 1.SG).

% To get a high-level idea of the amount of variation in label forms, we counted the number of unique lexical and grammatiical glosses...we include additional details in Appendix X.

% While it is impossible to completely normalize all IGT data---as doing so may interfere with the linguist's original intent in their annotation---we can at least attempt to map common variants of the same grammatical glosses to the same label...}

\section{Automatic IGT Generation}
% \ltcomment{Changed the section name from Experiments to Evaluation, moved the subsection on metrics to here as well}
Next, we evaluate the applicability of our dataset to the NLP task of automatic gloss prediction. We select the IGT data from the SIGMORPHON Shared Task on Interlinear Glossing \citep{ginn-etal-2023-findings} to use for evaluation and testing, as this data consistently adheres to a set of glossing conventions and has been evaluated on prior models. 

\subsection{Target Languages}
% \ltcomment{moved this section to evaluation as I think it would flow better there, but open to having it stay in section 3 if people think that would work better}

\label{sec:splits}
% \ltcomment{TODO: clarify splits re Graham's comments}
We reuse the train/eval/test splits for the seven languages from the SIGMORPHON Shared Task (\autoref{tab:datasplits}). We designate three languagues---Arapaho, Tsez, and Uspanteko---as \textit{in-domain languages}, which are included in the \textsc{GlossLM} corpus. We use Gitksan, Lezgi, Natugu, and Nyangbo as \textit{out-of-domain languages}, which are omitted from the corpus. All languages except Nyangbo include translations. 

% We evaluate the effectiveness of our pretrained model on automatic generation of IGT across the seven evaluation languages described in \S\ref{sec:splits}. 

% \subsection{Experimental Settings}
% For each language, we compare models under two settings:

% \mgcomment{If we're getting rid of segmented we can change this}

% \paragraph{Segmented} In the segmented setting, words in the input transcriptions are divided into morphemes with segmentation boundaries. Models must learn to classify each morpheme with an appropriate gloss, making this an easier setting (with SOTA models achieving 80-90\% accuracy). 

% \paragraph{Unsegmented} In the unsegmented setting, words in the input transcriptions are not segmented (but glosses are), requiring the model to jointly learn segmentation and glossing. Generally, this setting is far more challenging, with SOTA models achieving as low as 11\% accuracy on small datasets. \ltedit{As there is oftentimes far greater amounts of unsegmented transcribed data, automatically glossing from unsegmented data, though a more difficult task, can allow us to create magnitudes more gloss data in many languages.} \ltcomment{Is there a citation out there that might support this?} 

The shared task included two distinct settings:
\begin{itemize}
    \item In the \textit{open track}, the transcription lines were segmented into morphemes. This becomes a token classification task, and tends to be far easier, with SOTA models achieving 80-90\% accuracy (and even a naïve method that simply selects the most common gloss for each morpheme was very effective).
    \item In the \textit{closed track}, transcription lines were not segmented. This setting is far more challenging, as models must jointly learn to segment words and predict glosses, and the best models achieved as low as 11\% accuracy on small datasets. On the other hand, these models have the potential to be more valuable to documentation projects, where segmented text may not be available.
\end{itemize}

In our experiments, we focus on the unsegmented setting (closed track). However, the segmented data is also included in the \textsc{GlossLM} corpus, and could easily be used in future research.

\begin{table}[h]
\small
\centering
\resizebox{\columnwidth}{!}{%
    \begin{tabular}{lccccc}
    \toprule
     & & \multicolumn{3}{c}{Finetuning} \\
    \cmidrule(lr){3-5}
    Language & Pre-train & Train & Eval & Test \\
    \midrule
        Other languages & 198,121 & - & - & - \\
    \midrule
    \multicolumn{5}{l}{\textit{In-domain languages}} \\
    Arapaho (arp) & 39,132 & 39,132 & 4,892 & 4,892  \\
    Tsez (ddo) &  3,558 & 3,558 & 445 & 445 \\
    Uspanteko (usp) & 9,774 & 9,774 & 232 & 633 \\
    \midrule
    \multicolumn{5}{l}{\textit{Out-of-domain languages}} \\
    Gitksan (git) & - & 74 & 42 & 37 \\
    Lezgi (lez) & - & 705 & 88 & 87 \\
    Natugu (ntu) & - & 791 & 99 & 99 \\
    Nyangbo (nyb) & - & 2,100 & 263 & 263 \\

    \bottomrule
    \end{tabular}%
    }
    \caption{Number of total (unsegmented) pretrain (for in-domain languages), train, evaluation, and test examples for the target languages.}
    \label{tab:datasplits}
\end{table}

\subsection{Evaluation Metrics}
For evaluating predictions, we strip punctuation (except for within glosses).\footnote{Because of this post-processing, our results for baselines are slightly different than what original sources report.} We evaluate \textbf{morpheme accuracy}, which counts the number of correct morpheme glosses \textit{in the correct position}. Hence, if a gloss is incorrectly inserted or deleted, the subsequent glosses will be incorrect. We also evaluate \textbf{word accuracy}, which counts the number of entire correct word glosses. 

However, accuracy may sometimes be too strict of a measurement---especially for generative models---as minor character insertions/deletions in the label are penalized heavily. Thus, we also evaluate \textbf{chrF++} \citep{popovic-2015-chrf}, a character-level metric often used in machine translation. chrF++ measures the F1 score over character n-grams between the reference and predictions, and is robust to insertions and deletions, unlike accuracy.

\section{GlossLM Model}
\label{sec:model}
Using our IGT corpus described in \autoref{sec:corpus}, we train a single multilingual pretrained model for the glossing task that can be easily adapted to documentation projects, for both seen languages and unseen ones.

\subsection{Architecture} We use the ByT5 model, a multilingual pretrained model using the T5 architecture~\cite{raffel2020exploring}. ByT5 operates on byte-level inputs, as opposed to word or subword tokens, making it easily adaptable to different scripts and unseen languages. We use the ByT5-base model (582M parameters), pretrained on the mC4 dataset~\cite{xue-etal-2021-mt5}. We did not experiment with pretraining a randomly initialized model, as pretraining runs are expensive and we predict that the pretrained non-IGT base model serves as a better initialization.

\subsection{IGT Pretraining} We continually pretrain the ByT5 model on the \textsc{GlossLM} corpus described in \autoref{sec:corpus}. As we are evaluating on unsegmented IGT data, we omit segmented data for the evaluation languages from the pretraining corpus.\footnote{However, our corpus includes segmented IGT examples in other languages, which we do not evaluate.} We structure the glossing task as a text-to-text problem, training the model with examples formatted with the following prompt:\\

% The first (\textsc{GlossLM\textsubscript{all}}) uses all the training data in our dataset. Our second model (\textsc{GlossLM\textsubscript{unseg}}) excludes any segmented data in any of our test languages, simulating scenarios where a linguist does not have access to a corpus of segmented glossed data in their target language. Then, for each version of \textsc{GlossLM}, we finetune on the target language of interest, again using all data in the target language for \textsc{GlossLM\textsubscript{all}} and only unsegmented data in the target language for \textsc{GlossLM\textsubscript{unseg}}.

\noindent\texttt{\footnotesize
Provide the glosses for the following transcription in <lang>. \\ \\
Transcription in <lang>: <transcription> \\
Transcription segmented: <yes/no/unknown> \\
Translation in <metalang>: <translation> \\
Glosses: 
} \\

\noindent Models are trained to output the gloss line following the above prompt input. We include translations, which has been shown to provide benefits in gloss prediction~\citep{zhao_automatic_2020, ginn-etal-2023-findings}. For some data, a translation was not available ($\approx3\%$ of the training data), in which case the translation line is omitted.
We pretrain models using the hyperparameters given in \autoref{sec:hyperparam}. We did not conduct hyperparameter search, only tuning the batch size, to fit in our GPUs, and epochs and early stopping, to ensure convergence.

\subsection{Performance of Pretrained Model}
\label{sec:pretrained}
When training massively multilingual models, performance on individual languages can sometimes degrade in what is dubbed the "curse of multilinguality" \citep{conneau-etal-2020-unsupervised, chang2023multilinguality}. To investigate this issue, we evaluate our pretrained model on the in-domain languages without any additional finetuning. 

\begin{figure}[h]
    \centering
    \includegraphics[width=\linewidth]{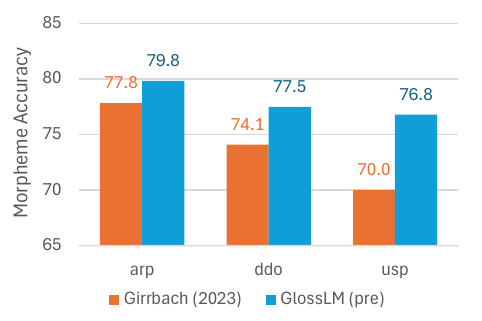}
    \caption{Comparison of our pretrained model and the SOTA \cite{girrbach-2023-sigmorphon} for in-domain languages on unsegmented data. Our model outperforms on all three languages. }
    \label{fig:pretrained-perf}
\end{figure}

We compare the performance of our pretrained model to the current SOTA, which is the second system from \citet{girrbach-2023-sigmorphon}, as shown in \autoref{fig:pretrained-perf}.  We find that the model outperforms the SOTA  across all three in-domain languages. This result gives little evidence to believe our model suffers from the curse of multilinguality, as it retains good performance across several languages.\footnote{These languages do make up large fractions of our pretraining corpus, so the model will almost certainly underperform on underrepresented languages.} 

Our pretrained model can be used for automated glossing across several languages, without needing to train and serve separate monolingual models. This could be valuable to real-world documentation projects, as we can serve a single pretrained model that can be used across projects, significantly reducing the barrier to using an automated system. 

The languages evaluated here are well-represented in the pretraining corpus, from Tsez (3.7k unsegmented examples, about 3\% of the total corpus) to Arapaho (39.1k examples, 21\%). A natural question is whether the model retains good performance on a language which occurs very rarely in the pretraining corpus. We simulate this scenario by adding a small amount of data to the pretraining corpus for two unseen languages: 1000 examples in Nyangbo and 500 in Natugu (less than 1\% of the total corpus). We evaluate on the unseen test split and observe 76.8\% and 55.0\% morpheme accuracy, respectively. These results indicate the model can still perform well on languages that are sparse in the pretraining corpus.

\section{Results}
% \subsection{\textsc{GlossLM}}
\label{sec:glosslm-experiments}

\subsection{Comparison with Baselines}
\label{sec:baselines}
\begin{figure*}[!bt]
    \centering
    \includegraphics[width=\linewidth]{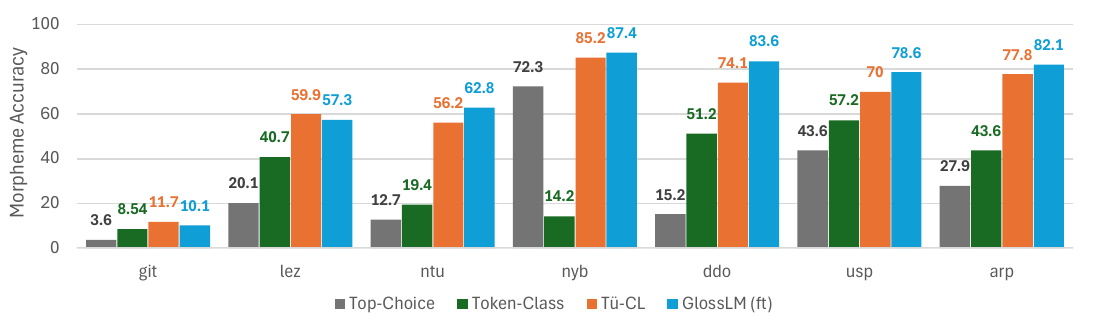}
    \caption{Morpheme accuracy for various systems. }
    \label{fig:sota}
\end{figure*}
After pretraining \textsc{GlossLM}, we train finetuned versions for each of the languages in the test set. We first describe our finetuning procedure and compare results of our finetuned models against baselines from~\citet{ginn2023sigmorphon} (\S\ref{sec:baselines}). Then, to further isolate the efficacy of pretraining, we compare the finetuned versions of \textsc{GlossLM} to a finetuned ByT5 model without the multilingual gloss pretraining (\S\ref{sec:byt5}). Finally, we explore whether training on a minimally normalized version of the data improves performance (\S\ref{sec:glosslm-norm}).

% We run three sets of experiments to measure the effectiveness of our approach, as well as one to explore the benefits of data normalization. First, to see whether the \textsc{GlossLM} model (without further finetuning) either benefits from multilingual gloss pretraining or suffers from the ``curse of multilinguality'', we compare the performance of the pretrained \textsc{GlossLM} to state-of-the-art monolingual models on languages in the GlossLM pretraining corpus (in-domain languages). To further isolate the efficacy of pretraining, we then compare finetuned versions of \textsc{GlossLM} on both in- and out-of-domain languages and compare the performance to a finetuned ByT5 model without the multilingual gloss pretraining. Then, we compare our finetuned models to SOTA systems. Finally, we explore whether training on a minimally normalized version of the data (as described in \S\ref{sec:normalizing}) improves performance.

As previously noted, we focus on the \textit{unsegmented} setting; for completeness, we provide full results for both segmented and unsegmented data in \autoref{sec:full-results}.

Finetuning can help align the model to a particular language or even a new unseen language. We finetune our \textsc{GlossLM} pretrained model on the training dataset for each language individually, and label these runs as \textsc{GlossLM\textsubscript{ft}}. We do this for both the in-domain languages, to focus the model on a single language, as well as the out-of-domain languages, allowing us to study the model's adaptation to unseen languages. 

Finetuning used the same parameters as pretraining, but with 100 training epochs and early stopping (patience 3, start epoch 15)\footnote{For Gitksan, due to the size of the training set, we set max epochs to 300 and patience to 15.}, and took anywhere from 20 minutes to one day for each language. Inference uses beam search with $n=3$ beams.

We compare the finetuned \textsc{GlossLM} models with three
baseline systems which include the state-of-the-art from prior work:

\begin{enumerate}
    \item \textbf{\textsc{Top-choice}} selects the most frequent label associated with each morpheme/word in the training data, and assigns ``\texttt{???}'' to unseen morphemes.

    \item \textbf{\textsc{Token-class}} treats the glossing task as a token classification problem, where the output vocabulary consists of the IGT morpheme or word-level labels. Each target language's data is used to train a language-specifc \textsc{Token-class} model, which uses the RoBERTa architecture with default hyperparameters without any additional pretraining~\cite{liu_roberta_2019}. This was used as the baseline model for the SIGMORPHON 2023 Shared Task on interlinear glossing~\cite{ginn2023sigmorphon}.

    \item \textbf{\textsc{T{\"u}-CL}}~\cite{girrbach-2023-tu-cl} uses straight-through gradient estimation \citep{bengio2013estimating} to induce latent, discrete segmentations of input texts, and predicts glosses using a multilayer perceptron.
\end{enumerate}

As shown in ~\autoref{fig:sota}, we find that our finetuned models outperform SOTA in all but two languages (Gitksan and Lezgi). The \textsc{T{\"u}-CL} model \citep{girrbach-2023-sigmorphon}, which is a close second and outperforms on Gitksan and Lezgi, uses explicit latent segmentations, which seems to be particularly beneficial for the very low-resource, unsegmented setting.

To illustrate common errors from the finetuned \textsc{GlossLM} moderls, we include examples of system outputs in~\autoref{tab:example_outputs}. When inspecting outputs, we observe that there are sometimes inconsistencies in the IGT labels where multiple interchangeable glosses are used for the same morpheme. While we try to account for a portion of grammatical gloss variations (\S\ref{sec:normalizing}, \S\ref{sec:glosslm-norm}), this is particularly an issue for lexical glosses (e.g. the Arapaho word \textit{'eeneisih'i} is glossed in the data as ``how.X.things.are.named'', ``how.s.o..is.named'', and ``how.things.are.named'').
% , which are harder to fix across individual corpora. These types of errors can be somewhat accounted for when using \textsc{chrF++} as an evaluation metric, but another issue that may more greatly impact generative models---especially if it learns to use translations effectively---is if the model outputs a \textit{semantically} similar label that is dissimilar in form; for example, outputting ``PREP'' (preposition label) instead of ``en'' (``in'' in Spanish, which is a preposition)
We also find outputs that include lexical items present in the translation that are not included in the gold gloss, indicating that the model may rely too heavily on translations when predicting lexical glosses in certain cases.

\subsection{Comparison with Finetuned ByT5}
\label{sec:byt5}
\begin{figure}[h]
    \centering
    \includegraphics[width=\linewidth]{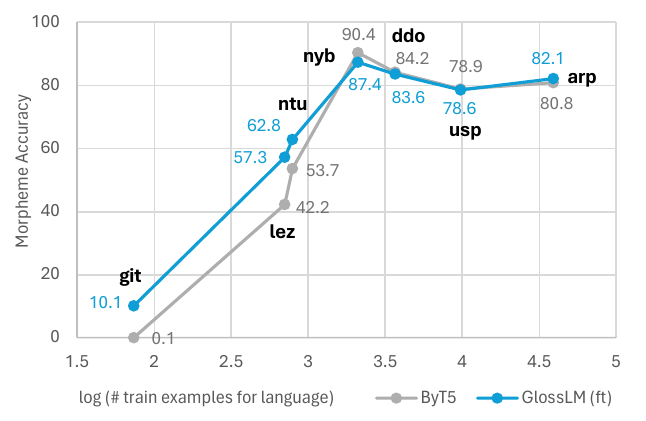}
    \caption{Performance after monolingual finetuning, comparing a standard pretrained ByT5 with a continually pretrained \textsc{GlossLM} model. The x-axis uses the log (base 10) of the number of training examples in a given language, for readability.}
    \label{fig:finetuned-perf}
\end{figure}

To directly assess the impact of pretraining on performance, we finetune ByT5 models on each language in the test set with the same configuration as for the finetuned \textsc{GlossLM} models.
We then compare the performance of our models (which have undergone both multilingual gloss pretraining and finetuning) with analogous ByT5 models (without multilingual gloss pretraining), as shown in \autoref{fig:finetuned-perf}.

We observe mixed results, which are largely dependent on the size of the training corpus. For languages with less training data (Gitksan, Lezgi, Natugu) the \textsc{GlossLM\textsubscript{ft}} model outperforms the finetuned ByT5 model (by 10.0, 15.1, and 9.1 points respectively). In the case of Gitksan, the finetuned ByT5 model is completely unable to produce well-formatted output (likely due to the tiny training corpus) while the \textsc{GlossLM\textsubscript{ft}} model does not struggle with this as much. A possible explanation is that even if there are no similar languages in the pretraining corpus, the \textsc{GlossLM\textsubscript{ft}} can leverage knowledge about IGT formatting from unrelated languages. 

For Lezgi---which shows the greatest improvements from pretraining with the \textsc{GlossLM} corpus---a qualitative analysis of examples with the greatest morpheme error rate between finetuned ByT5 and \textsc{GlossLM} reveals that there are regular error patterns that are fixed with continual pretraining. For example, the finetuned ByT5 model often outputs \textsc{AOC} instead of \textsc{AOR} and \textsc{OLB} instead of \textsc{ERG}, whereas the finetuned \textsc{GlossLM} gets these correct. We include examples of these outputs in~\autoref{tab:lez_example_outputs}.

However, with enough data (starting with Nyangbo, 2100 examples) the two approaches achieve nearly identical performance. This is an unsurprising result, indicating that large amounts of high-quality monolingual data overshadow any benefits from crosslingual transfer. Furthermore, we note that benefits of multilingual gloss pretraining shown may not be unique to the T5 architecture---while we only experiment with ByT5, our pretraining strategy could be applied to other architectures.

\subsection{Effect of Gloss Normalization}
\label{sec:glosslm-norm}
Finally, we experiment with pretraining and finetuning on a minimally normalized version of the dataset, where the 200 most frequent grammatical labels are mapped to a set of standard labels.

We repeat the same pretraining and finetuning process as before. When comparing the performance of the pretrained model before finetuning on in-domain languages, we find minimal differences compared to the results in \S\ref{sec:pretrained}. 

\begin{figure}
    \centering
    \includegraphics[width=\linewidth]{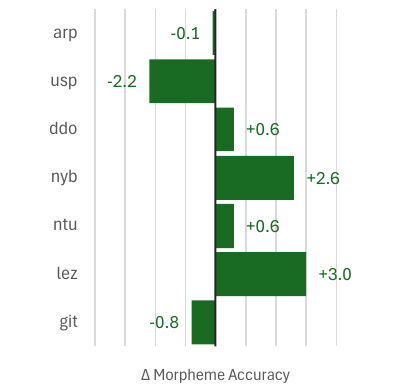}
    \caption{Change in morpheme accuracy after normalizing glosses to the UniMorph schema and finetuning \textsc{GlossLM}. We observe small improvements for several languages, but worse performance in two cases.}
    \label{fig:normalization}
\end{figure}

We report the change in morpheme accuracy after normalizing in \autoref{fig:normalization}. We observe mixed results. Some languages (Arapaho, Uspanteko, Gitksan) show worse or equivalent performance. Others (Tsez, Nyangbo, Natugu, and Lezgi) show small to moderate improvements, with Lezgi achieving the largest improvement of 3.0 percentage points. Thus, we found that normalization was most helpful when finetuning the pretrained model on unseen languages with a low-to-moderate amount of training data.

\section{Related Works}
\paragraph{Automatic Interlinear Glossing} Recent research has explored various methods for generating IGT, including rule-based methods \citep{bender2014learning}, active learning \citep{palmer_computational_2010, palmer2009evaluating}, conditional random fields \citep{moeller_automatic_2018, mcmillan-major_automating_2020}, and neural models \citep{moeller_automatic_2018, zhao_automatic_2020}. The 2023 SIGMORPHON Shared Task \citep{ginn-etal-2023-findings} compared a number of highly-effective IGT generation systems, including ensembled LSTMs \citep{coates-2023-ensembled}, straight-through gradient estimation \citep{girrbach-2023-sigmorphon}, CRF-neural systems \citep{okabe-yvon-2023-lisn}, and BiLSTM encoders \citep{cross-etal-2023-glossy}. 

In particular, this work is inspired by \citet{he-etal-2023-sigmorefun}, which pretrains ByT5 models on the ODIN corpus, and \citet{okabe-yvon-2023-towards}, which pretrains a CRF model on the IMTVault corpus. However, neither explore using a pretraining corpus as large as ours, nor do they evaluate on unsegmented text. Furthermore, neither of these studies find significant benefits to using pretraining corpora, while we observe benefits under certain conditions.

\paragraph{Large Multilingual Pretrained Models}
Prior work has shown that large, massively multilingual pretrained language models can boost performance across low- and high-resource languages on a variety of tasks. These include encoder-decoder models trained with the masked language modeling objective~\cite{pires-etal-2019-multilingual, conneau-etal-2020-unsupervised} and the span corruption objective~\cite{xue-etal-2021-mt5, xue-etal-2022-byt5}, as well as decoder-only language models~\cite{workshop2023bloom, shliazhko2024mgpt}. Work such as~\citet{wang-etal-2020-extending}, \citet{adelani-etal-2022-masakhaner}, and \citet{adelani-etal-2022-thousand} has shown that continual pretraining and/or finetuning large multilingual models is an effective method for tasks like low-resource language translation and named entity recognition.

\section{Conclusion}
High-quality language documentation involves an incredible amount of effort. We compile, normalize, and release the largest corpus of multilingual IGT data, enabling future research in linguistics, NLP, and documentation. Furthermore, we demonstrate the applicability of our corpus by pretraining a multilingual neural model for automatic generation of IGT. We finetune the model on monolingual corpora, showing benefits on low-resource languages due to multilingual pretraining. In five out of seven languages, we achieve a new SOTA on automatic IGT generation.

% Automatic generation of interlinear glossed text has great potential to aid documentation projects, but limited monolingual data limits effectiveness. In this research, we attempt to overcome these limitations using crosslingual transfer, compiling the largest multilingual IGT dataset and pretraining massively multilingual glossing models. Our pretrained model is highly competitive for langauges in its pretraining corpus, and finetuned models achieve a new SOTA in five out of seven languages. Furthermore, we observe clear benefits from the multilingual pretraining after finetuning for languages with small training corpora. We make our dataset and models available for future research, in hopes that they will aid future NLP research and benefit documentation projects.

\section*{Limitations}
% \apmargincomment{Shouldn't Limitations and Ethics sections be un-numbered? Check submission guidelines}
Our work evaluates the effectiveness of massively multilingual pretraining on seven IGT datasets in different languages using the ByT5 architecture. We did not experiment with pretraining on other architectures, which may show different results. While we believe the selected evaluation languages cover a diverse set of features and dataset sizes, other languages may show better or worse results.

Our pretraining corpus consists of all  IGT data we were able to find and utilize. As such, it is not evenly distributed among languages, over-representing a few languages with large language documentation efforts. Thus, models pretrained on the corpus will perform better on these and similar languages.

The only hyperparameter optimization we performed was finding a batch size that fit our GPUs and tuning epochs and early stopping in order to ensure convergence. We did not conduct hyperparameter search over other parameters such as learning rate or optimizer, architecture parameters, or dataset splits.

When evaluating predictions, we ignored punctuation (as our primary concern was gloss accuracy). Certain models may perform better or worse at outputting proper punctuation format, which could be a concern for certain applications. 

Finally, it has been demonstrated that IGT generation models are often not robust to domain shift, compared with human annotators \citep{ginn-palmer-2023-robust}. Our models will likely have impacted performance for out-of-domain texts, such as highly technical or domain-specific language.

\section*{Ethics Statement}
We hope this work can aid in the struggle against language extinction. However, language documentation, preservation, and revitalization require far more than generating IGT, and we should be careful not to understate the difficulty of these efforts. We utilize datasets produced by the painstaking effort of language documenters and speakers, and strive to treat the corpora as human artifacts, not just data to be consumed. 

We hope our research can aid documentary linguists through automated gloss prediction. However, we caution against using these systems without human collaboration, as they can introduce error and miss novel linguistic insights. There is some risk of these systems being used to replace human annotators, which we strongly oppose. 

While we try to train only the necessary models for our experiments, training large machine learning models carries an environmental cost \citep{bender_dangers_2021, strubell2020energy}. 

We do not evaluate our corpus for bias (racial, gender, etc) or inclusive language, and it's possible that our models can carry some of these biases. 

Finally, NLP work that involves Indigenous and endangered languages has historically been plagued by colonialist approaches to data use and technology development \citep{schwartz-2022-primum}. The large IGT datasets for endangered languages (Arapaho, Guarani, Uspanteko) were collected in collaboration with native communities, and our work is in accordance with the agreements for their usage. 

\section*{Acknowledgements}
We would like to thank the three anonymous reviewers and meta-reviewer for their helpful feedback on this work.
Parts of this work were supported by the National Science
Foundation under Grant No. 2149404, “CAREER: From One Language to Another” and Grant No. 2211951, "From Acoustic Signal to Morphosyntactic Analysis in One End-to-End Neural System." Any opinions, findings, and conclusions or recommendations expressed in this material are those of the authors and do not necessarily reflect the views of the National
Science Foundation.

% Bibliography entries for the entire Anthology, followed by custom entries
\bibliography{anthology,custom}
% Custom bibliography entries only
% \bibliography{custom}

\appendix
\newpage

\begin{figure*}[!tb]
  \centering
  \vspace{40pt}
   \includegraphics[width=\textwidth]{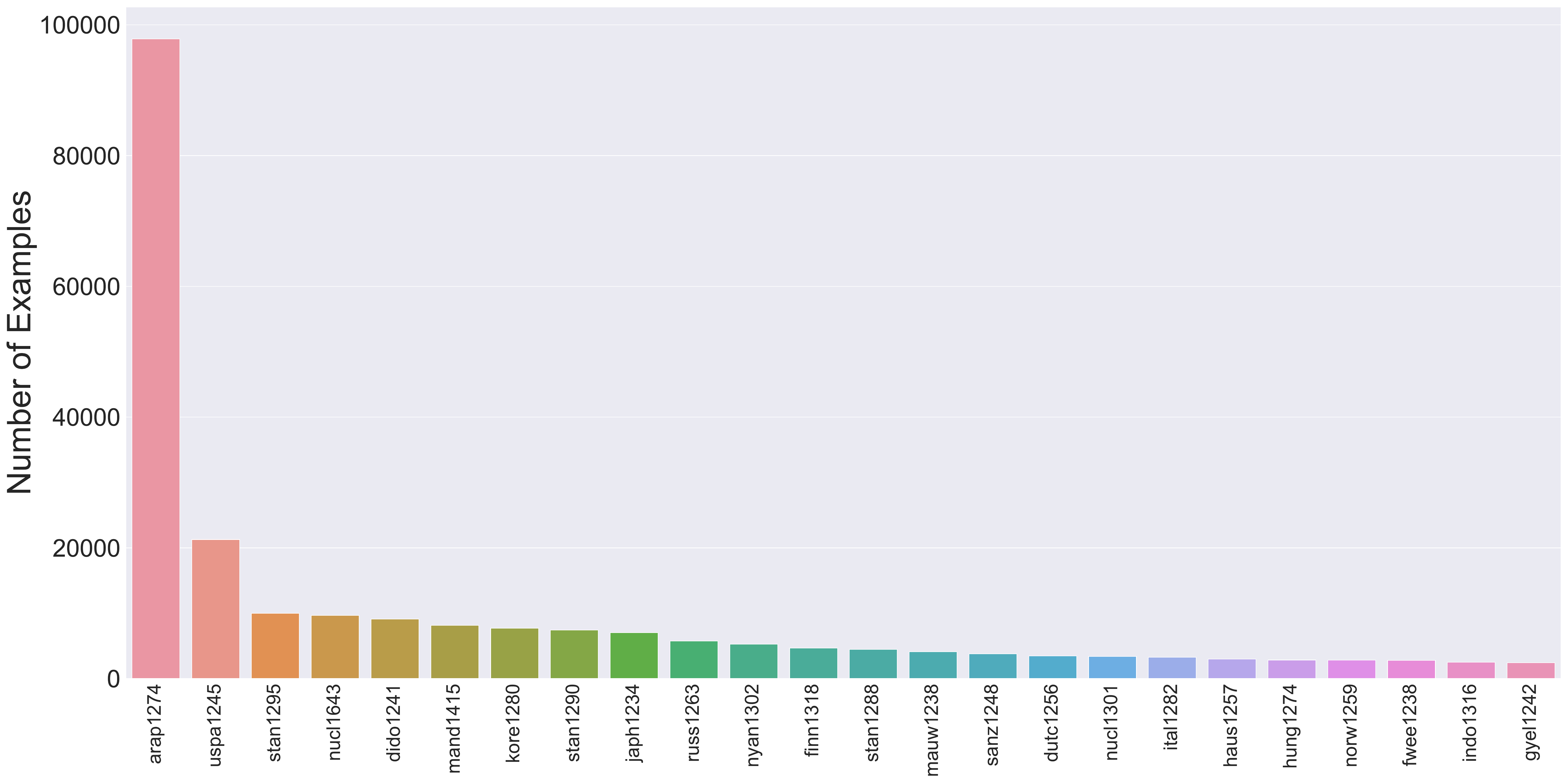}
  \caption{Counts per language. We only show languages with at least 2k samples present in the dataset. Arapaho (arap1274) is by far the most represented language in our data, followed by Uspanteko (uspa1245). Both languages are part of the SIGMORPHON Shared Task dataset.} 
  \vspace{40pt}
\label{fig:lang_counts}
\end{figure*}

\begin{figure*}[!tbh]
  \centering
   \includegraphics[width=\textwidth]{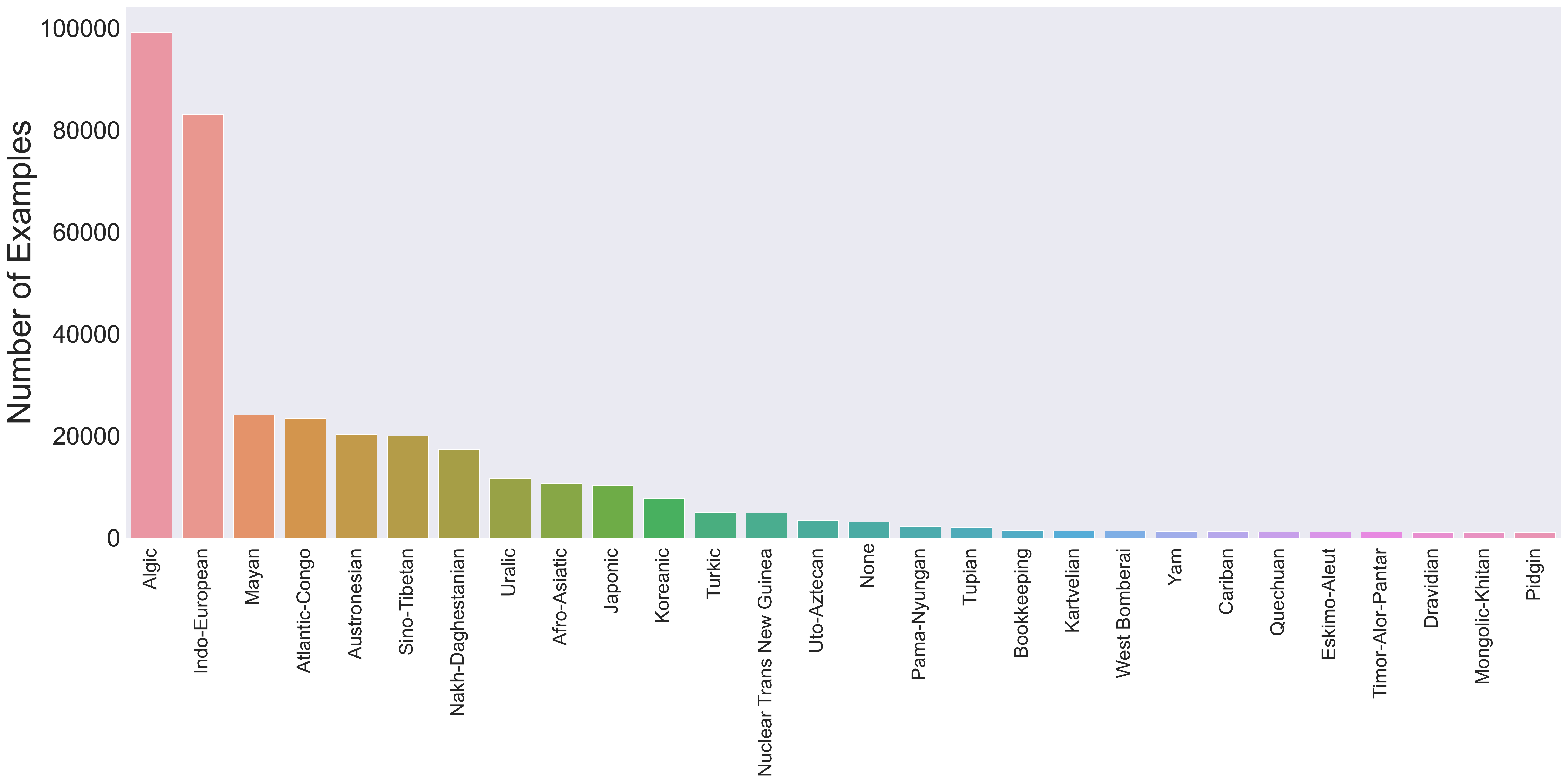}
  \caption{Counts per language family in our dataset. We only show language families with at least 1k samples present in the dataset. Language isolates and languages without recorded families in Glottolog~\cite{hammarstrom2023glottolog} are categorized under ``None''.}
  \vspace{40pt}
\label{fig:lang_family}
\end{figure*}

\section{Language Distribution}
\label{sec:langdist}

\noindent \autoref{fig:lang_counts} and~\autoref{fig:lang_family} display the number of examples per language and language family on a portion of our dataset.

\section{Grambank Typological Analysis}
\label{sec:grambank}
Along with number of languages, we would also like to measure whether the distribution of typological features in our dataset is reflective of the diversity of features in the world. We use the Grambank typological database~\cite{grambank_release} as a standard against which to judge the typological diversity of our dataset. Grambank covers over 2430 languages, with up to 195 features (e.g., "What is the order of numeral and noun in the NP?") per language. The values of the features comprise a vector for each language. 

43\% of our languages are found in Grambank, amounting to 72\% coverage over all training instances. However, Grambank does not have complete feature vectors for all languages. Using the method described by~\citet{grambank_release}, we imputed missing feature values (9.7\% of all features), resulting in complete feature vectors of size 161, as we only accept features that are defined for at least $64\%$ of our dataset (as to balance dataset coverage as well as feature coverage). 

We then create an average feature vector for our dataset by averaging the feature vectors of the the languages present in Grambank (weighting the average based on the number of training instances in each language) and compare this to the average of feature vectors for \textit{all} languages in Grambank. We find a cosine similarity of 0.92 between the two vectors. In comparison, the language with the greatest similarity to the average Grambank vector in our data has a cosine similarity of 0.81. We report additional details of our methods and analysis below.

% In order to obtain a rough sketch of typological coverage, we compute an average Grambank \cite{grambank_release} feature vector over all languages in our training set, weighted by the number of training instances in each language. Though not all our training languages are represented in Grambank, we find matching feature vectors comprising of 161 features for 43.71\% of our languages, amounting to 71.94\% coverage over all training instances. The full set of average features is provided in \autoref{sec:grambank}. Furthermore, not all of our languages are defined over the full Grambank feature space, so we impute 9.7\% of all features. A more in-depth description of the imputation process is included in \autoref{sec:grambank}. We compare our average feature vector to an expected feature vector obtained by averaging over all 2,430 languages covered by Grambank, and find a cosine similarity of 0.92, which we take an indicator that our training data represents a realistic typological distribution. We report the 5 features with an average value most distant from the expected value in \autoref{sec:grambank}.

\subsection{Imputation Details}

We adapt the imputation procedure described in \cite{grambank_release} and follow the steps below. Thresholds were chosen to maximize the language coverage while keeping the imputed values below 10\%.
\begin{itemize}
   \item Removed languages that had > 36\% missing data out of the dataset
    \item Removed features that had >36\% missing data among the remaining languages
    \item Binarized the multistate values 
    \item Removed all but one dialect for each language according 
    \item Imputed missing values with iterated random forest with MissForest\footnote{https://rpubs.com/lmorgan95/MissForest}
\end{itemize}

\subsection{Underrepresented Features}
\begin{table}[h]
\small
\centering
\resizebox{\columnwidth}{!}{%
    \begin{tabular}{clc}
    \toprule
    \textbf{ID} & \textbf{Feature} & \textbf{Average Value} \\
    \midrule
    \multirow{2}{*}{GB024b} & Does the numeral always follow the noun & \multirow{2}{*}{0.468} \\ 
    & in the NP? & \\
    \midrule
    \multirow{2}{*}{GB193b} & Are most adnominal property words placed & \multirow{2}{*}{0.536} \\ 
    & after nouns? & \\
    \midrule
    \multirow{2}{*}{GB025b} & In the pragmatically unmarked order, does the & \multirow{2}{*}{0.425} \\ 
    & adnominal demonstrative follow the noun? & \\
    \midrule
    GB118 & Are there serial verb constructions? & 0.406 \\ 
    \midrule
    \multirow{2}{*}{GB130a} & Is the order in intransitive clauses with & \multirow{2}{*}{0.382} \\ 
    & a full nominal subject consistently SV? & \\
    \bottomrule
    \end{tabular}%
    }
    \caption{Most underrepresented Grambank features in training data with their average value (from imputed vectors).}
    \label{tab:grambank_features}
\end{table}

In \autoref{tab:grambank_features} we report the top five features for which the average representation in our training data is most distant from the expected value (as determined by averaging feature values across all languages in Grambank). 

\subsection{Averaged Feature Vector}
For reference, we include the vector set of 161 average Grambank features over all training languages weighted by the number in \autoref{tab:avg_gram}.

\section{Training Hyperparameters}
\label{sec:hyperparam}
Training the \textsc{GlossLM\textsubscript{all}} and \textsc{GlossLM\textsubscript{unseg}} models used A6000 and A100 GPUs, and took around 5 days per run. We list the hyperparameters used in~\autoref{tab:traininghyper}.

\begin{table}[!htb]
\small
    \centering
    % \begin{adjustbox}{width=\linewidth}
    \begin{tabular}{l c}
        \toprule
        Parameter & Value \\
        \midrule
        Optimizer & Adafactor  \\
        Initial LR & 5e-5 \\
        Weight decay & 0.01 \\
        Batch size & 2 \\
        Gradient accumulation steps & 64 \\
        Epochs & 13 \\
        \bottomrule
    \end{tabular}
    % \end{adjustbox}
    \caption{Pretraining Hyperparameters}
    \label{tab:traininghyper}
\end{table}

\begin{table*}[!htb]
\small
\centering
\resizebox{\textwidth}{!}{
    \def\arraystretch{1.2}
    \begin{tabular}{l | c c c c | c c c c c}
    \toprule
    \multirow{3}{*}{\textbf{Model}} & \multicolumn{9}{c}{Morpheme accuracy / Word accuracy (Segmented)} \\
    & \multicolumn{4}{c|}{\textit{In-domain}} & \multicolumn{5}{c}{\textit{Out-of-domain}} \\
    & \textbf{arp} &  \textbf{ddo} & \textbf{usp} & \textit{Avg} & \textbf{git} & \textbf{lez} & \textbf{ntu} & \textbf{nyb} & \textit{Avg} \\
    \midrule
    \textsc{Top-choice} & 83.2 / 74.0 & 78.5 / 64.4 & 79.7 / 72.9 & 80.5 / 70.4 & 51.1 / 29.7 & 62.2 / 54.4 & 78.4 / 68.1 & 72.5 / 63.8 & 66.1 / 54.0 \\
    \textsc{Token-class} & 90.4 / 84.2 & 86.4 / 76.5 & 82.5 / 76.5 & 86.4 / 79.1 &  25.3 / 16.4 & 50.2 / 38.8 & 62.0 / 54.3 & 88.8 / 84.4 & 56.6 / 48.5 \\
    \textsc{T{\"u}-CL} & \textbf{90.7} / 84.6 & 91.2 / 85.1 & 84.5 / 78.5 & 88.8 / 82.7 &  50.2 / 26.6 & \textbf{84.9} / \textbf{77.6} & \textbf{91.7} / 86.0 & \textbf{91.4} / \textbf{87.9} & \textbf{79.6} / \textbf{69.5} \\
    \textsc{CRF} & 90.4 / 84.2 & 91.9 / 85.6 & 84.4 / 78.9 & 88.9 / 82.9 &  \textbf{52.4} / 33.6 & 84.7 / 77.5 & 91.1 / \textbf{86.6} & 88.8 / 84.4 & 79.3 / 70.5 \\
    \textsc{SMF} & 80.1 / 79.4 & 78.2 / 82.8 & 73.2 / 75.7 & 77.2 / 79.3 & 12.7 / 20.6 & 47.8 / 56.4 & 64.0 / 75.7 & 85.4 / 82.7 & 52.5 / 58.9 \\
    \textsc{ByT5\textsubscript{all}} & 88.7 / 83.2 & \textbf{93.5} / \textbf{89.9} & 86.3 / 82.7 & 89.5 / 85.3 & 2.2 / 3.6 &  72.5 / 69.7 & 83.4 / 82.2 & 90.7 / 89.2 & 62.2 / 61.2 \\
    \midrule
    \textsc{GlossLM\textsubscript{all, pre}} & 89.3 / 84.2 & 91.7 / 88.3 & 84.1 / 81.0 & 88.4 / 84.5 & 3.6 / 9.1 & 3.6 / 1.8 & 4.9 / 9.8 & 2.9 / 3.0 & 3.8 / 5.9 \\
    \textsc{GlossLM\textsubscript{all, ft}} & 90.1 / \textbf{85.0} & 92.8 / 89.3 & \textbf{86.4} / \textbf{84.5} & \textbf{89.8} / \textbf{86.3} &  28.9 / \textbf{34.9} & 74.7 / 71.3 & 86.0 / 81.5 & 90.7 / 87.7 & 70.1 / 68.9 \\
    \bottomrule
    \end{tabular}
}

\resizebox{\textwidth}{!}{
    \def\arraystretch{1.2}
    \begin{tabular}{l | c c c c | c c c c c}
    \toprule
    \multirow{3}{*}{\textbf{Model}} & \multicolumn{9}{c}{Morpheme accuracy / Word accuracy (Unsegmented)} \\
    & \multicolumn{4}{c|}{\textit{In-domain}} & \multicolumn{5}{c}{\textit{Out-of-domain}} \\
    & \textbf{arp} &  \textbf{ddo} & \textbf{usp} & \textit{Avg} & \textbf{git} & \textbf{lez} & \textbf{ntu} & \textbf{nyb} & \textit{Avg} \\
    \midrule
    \textsc{Top-choice} & 27.9 / 56.9 & 15.2 / 64.1 & 43.6 / 60.4 & 28.9 / 60.5 & 3.6 / 16.9 & 20.1 / 58.2 & 12.7 / 55.1 & 72.3 / 76.7 & 27.2 / 51.7  \\
    \textsc{Token-class} & 43.6 / 69.9 & 51.2 / 74.3 & 57.2 / 72.1 & 50.7 / 72.1 &  8.54 / 16.9 & 40.7 / 45.5 & 19.4 / 48.2 & 14.2 / 5.96 & 20.7 / 29.1  \\
    \textsc{T{\"u}-CL} & 77.8 / 77.5 & 74.1 / 80.4 & 70.0 / 73.4 & 74.0 / 77.1 &  \textbf{11.7} / 21.1 & 59.9 / \textbf{71.8} & 56.2 / 78.0 & 85.2 / 85.0 & 53.3 / 64.0 \\
    \textsc{ByT5\textsubscript{unseg}} & 80.8 / 79.7 & \textbf{84.2} / 87.4 & \textbf{78.9} / \textbf{82.5} & 81.3 / 83.2 & 0.1 / 0.3 & 42.2 / 53.4 & 53.7 / 71.0 & \textbf{90.4} / \textbf{88.4} & 46.6 / 53.3  \\
    \midrule
    \textsc{GlossLM\textsubscript{unseg, pre}} & 79.8 / 79.2 & 77.5 / 82.8 & 76.8 / 80.8 & 78.0 / 80.9 & 2.3 / 3.5 & 1.5 / 1.3 & 4.1 / 9.6 & 1.6 / 2.9 & 2.4 / 4.3 \\
    \textsc{GlossLM\textsubscript{unseg, ft}} & \textbf{82.1} / \textbf{81.5} & 83.6 / 87.3 & 78.6 / 81.0 & \textbf{81.4} / \textbf{83.3} &  10.1 / \textbf{28.4}  & 57.3 / 64.9 & 62.8 / \textbf{78.9} & 87.4 / 86.2 & 54.4 / \textbf{64.6} \\
    \midrule
    \textsc{GlossLM-norm\textsubscript{unseg, pre}} & 79.6 / 80.0 & 79.6 / 83.2 & 74.8 / 76.6 & 78.0 / 79.9 & 2.2 / 7.8 & 2.6 / 1.8 & 2.9 / 9.7 & 1.0 / 2.5 & 2.2 / 5.45 \\
    \textsc{GlossLM-norm\textsubscript{unseg, ft}} & 82.0 / \textbf{81.5} & \textbf{84.2} / \textbf{87.8} & 76.4 / 79.2 & 80.8 / 82.8 & 9.3 / 16.4 & \textbf{60.3} / 67.8 & \textbf{63.4} / 76.6 & 90.0 / 87.6 & \textbf{55.8} / 62.1 \\
    \bottomrule
    \end{tabular}%
}

\caption{Morpheme- and word-level accuracy of various systems on segmented (top) and unsegmented (bottom) text. Best performance per language in each setting the table is \textbf{bolded}. \textsc{GlossLM\textsubscript{all, pre}} refers to performance using the pretrained \textsc{GlossLM} directly, while \textsc{GlossLM\textsubscript{all, ft}} refers to performance after fine-tuning the pretrained model on the specific language.} 
\label{tab:acc}
\end{table*}
\begin{table*}[!htb]
\small
\centering
\resizebox{0.8\textwidth}{!}{
    \begin{tabular}{l | c c c c | c c c c c}
    \toprule
    \multirow{3}{*}{\textbf{Model}} & \multicolumn{9}{c}{chrF++ Score (Segmented)} \\
    & \multicolumn{4}{c|}{\textit{In-domain}} & \multicolumn{5}{c}{\textit{Out-of-domain}} \\
    & \textbf{arp} &  \textbf{ddo} & \textbf{usp} & \textit{Avg} & \textbf{git} & \textbf{lez} & \textbf{ntu} & \textbf{nyb} & \textit{Avg} \\
    \midrule
    \textsc{Top-choice} & 75.0 & 71.9 & 71.4 & 72.8 & 33.7 & 69.7 & 74.5 & 62.2 & 60.0 \\
    \textsc{Token-class} & 84.2 & 81.8 & 75.3 & 80.4 & 25.6 & 52.1 & 65.4 & 84.3 & 56.9 \\
    \textsc{T{\"u}-CL} & 85.2 & 88.4 & 77.7 & 83.8 & 34.8 & 78.9 & 87.9 & 87.7 & 72.3 \\
    \textsc{CRF} & 84.2 & 88.2 & 79.4 & 83.9 & 40.9 & \textbf{79.2} & \textbf{88.4} & 84.3 & \textbf{73.2} \\
    \textsc{SMF} & 80.7 & 86.6 & 76.3 & 81.2 & 28.0 & 62.0 & 78.8 & 82.2 & 62.6 \\
    \textsc{ByT5\textsubscript{all}} & 84.2 & \textbf{91.8} & 84.6 & 86.9 & 9.07 & 74.9 & 85.1 & \textbf{88.8} & 64.5  \\
    \midrule
    \textsc{GlossLM\textsubscript{all, pre}} & 85.2 & 90.8 & 83.1 & 86.4 & 21.5 & 14.1 & 18.2 & 8.8 & 15.7 \\
    \textsc{GlossLM\textsubscript{all, ft}} & \textbf{86.3} & 91.5 & \textbf{85.9} & \textbf{87.9} & \textbf{43.1} & 75.0 & 84.1 & 87.4 & 72.4 \\
    \bottomrule
    \end{tabular}
}

\vspace{2pt}

\resizebox{0.8\textwidth}{!}{
    \begin{tabular}{l | c c c c | c c c c c}
    \toprule
    \multirow{3}{*}{\textbf{Model}} & \multicolumn{9}{c}{chrF++ Score (Unsegmented)} \\
    & \multicolumn{4}{c|}{\textit{In-domain}} & \multicolumn{5}{c}{\textit{Out-of-domain}} \\
    & \textbf{arp} &  \textbf{ddo} & \textbf{usp} & \textit{Avg} & \textbf{git} & \textbf{lez} & \textbf{ntu} & \textbf{nyb} & \textit{Avg} \\
    \midrule
    \textsc{Top-choice} & 44.0 & 63.5 & 55.1 & 54.2 &  8.4 & 51.6 & 40.5 & 74.0  & 43.6 \\
    \textsc{Token-class} & 56.2 & 72.9 & 65.3 & 64.8 & 18.8 & 56.4 & 45.1 & 18.8 & 34.8  \\
    \textsc{T{\"u}-CL} & 77.6 & 84.6 & 72.5 & 78.2 & 23.0 & \textbf{71.5} & 78.6 & 84.1 & 64.3 \\
    \textsc{ByT5\textsubscript{unseg}} & 80.7 & \textbf{90.0} & \textbf{83.0} & 84.6 & 7.6 & 59.6 & 77.0 & \textbf{88.4} & 58.2 \\
    \midrule
    \textsc{GlossLM\textsubscript{unseg, pre}} & 80.5 & 86.8 & 81.0 & 82.8 & 19.4 & 13.3 & 16.3 & 8.1 & 14.3 \\
    \textsc{GlossLM\textsubscript{unseg, ft}} & \textbf{82.9} & 89.8 & 81.7 & \textbf{84.8} & \textbf{34.9} & 68.8 & \textbf{80.7} & 85.5 & \textbf{67.5}  \\
    \midrule
    \textsc{GlossLM-norm\textsubscript{unseg, pre}} & 80.3 & 86.6 & 78.7 & 81.9 & 19.5 & 14.3 & 16.6 & 7.4 & 14.5 \\
    \textsc{GlossLM-norm\textsubscript{unseg, ft}} & 82.6 & 89.9 & 81.1 & 84.5 & 29.1 & 70.9 & 79.0 & 87.5 & 66.6 \\
    \bottomrule
    \end{tabular}
}
\caption{\textsc{chrF++} scores of various systems on segmented (top) and unsegmented (bottom) data. Best performance per language in each setting the table is \textbf{bolded}. \textsc{GlossLM\textsubscript{all, pre}} refers to performance using the pretrained \textsc{GlossLM} directly, while \textsc{GlossLM\textsubscript{all, ft}} refers to performance after finetuning on the specific language.} 
\label{tab:chrf}
\vspace{40pt}
\end{table*}

\section{Full Results}
We provide full results for accuracy and chrF++ scores in \autoref{tab:acc} and \autoref{tab:chrf}
\label{sec:full-results}. For our normalization experiments, we only trained and tested on unsegmented data for the target languages.

% \section{chrF++ Scores}
% \label{sec:chrf}
% We provide chrF++ scores for various systems in \autoref{tab:chrf}. Broadly, we found the results supported the same trends as the accuracy measurements, although the magnitude of the differences varied.

\section{In- vs out-of-vocabulary errors}
\begin{table}[!ht]
\small
\centering
\resizebox{\columnwidth}{!}{%
    \begin{tabular}{lccccccc}
    \toprule
     & \multicolumn{3}{c}{\textit{In-domain}} & \multicolumn{4}{c}{\textit{Out-of-domain}} \\
     \cmidrule(lr){2-4} \cmidrule(lr){5-8}
     & arp & ddo & usp & git & lez & ntu & nyb \\
     \midrule
     \midrule
     \% OOV & 30.0 & 15.6 & 20.0 & 78.1 & 27.3 & 27.6 & 8.42 \\
     \midrule
     IV & 96.2 & 92.3 & 91.4 & 66.7 & 85.4 & 91.3 & 92.8 \\
     OOV & 55.7 & 71.7 & 57.1 & 26.0 & 33.9 & 55.7 & 32.6 \\
     \midrule
     \midrule
     \% OOV & 30.0 & 18.7 & 21.4 & 80.5 & 25.5 & 28.9 & 9.27 \\
     \midrule
     IV & 95.3 & 91.4 & 89.1 & 60.0 & 81.6 & 91.3 & 91.7 \\
     OOV & 50.1 & 69.7 & 50.9 & 20.7 & 16.4 & 48.3 & 30.6 \\
    \bottomrule
    \end{tabular}%
    }
    \caption{Percent of words that are out-of-vocab in the test split for each language along with in- versus out-of-vocabulary accuracy at the word level. Top is the segmented setting (\textsc{GlossLM\textsubscript{all, ft}}), bottom is unsegmented (\textsc{GlossLM\textsubscript{unseg, ft}}).}
    \label{tab:iv_oov_acc}
    % \vspace{-10pt}
\end{table}
\begin{table}[!h]
\small
\centering
% \begin{minipage}{0.9\columnwidth}
    \begin{tabular}{lcc}
    \toprule
    Language & Morpheme \%OOV \\
    \midrule
    arp & 3.6 \\
    ddo & 41.2 \\
    usp & 4.9 \\
    git & 2.8 \\
    lez & 1.1 \\
    ntu & 0.5 \\
    nyb & 5.3 \\
    \bottomrule
    \end{tabular}%
    \caption{Percent of out-of-vocabulary morphemes in the test split.}
    % \end{minipage}
    \label{tab:morph_oov}
\end{table}
\begin{table}[!htb]
\small
% \begin{minipage}{0.9\columnwidth}
    \centering
    \begin{tabular}{lcc}
    \toprule
    Language & OOV Token Recall \\
    \midrule
    arp & 49.87 \\
    ddo & 44.13 \\
    usp & 44.89 \\
    git & 58.21 \\
    lez & 71.66 \\
    ntu & 40.14 \\
    \bottomrule
    \end{tabular}%
    \caption{Percent of lexical glosses present in the translation in the test split. Nyangbo examples do not include translations.}
    % \end{minipage}
    \label{tab:token_translation_recall}
\end{table}
We report word-level accuracy for our finetuned \textsc{GlossLM} models, indicating whether the transcribed word is in- or out-of-vocabulary in \autoref{tab:iv_oov_acc}, as well as the percent of OOV words in the test set. The OOV rate between segmented and unsegmented may vary slightly, as mappings between segmented and unsegmented forms are not necessarily one-to-one. We consider a word to be in-vocabulary if the form of the word in the transcription \textit{and} its corresponding gold label in the gloss co-occur in the training data. We also include morpheme-OOV rates and statistics on lexical gloss overlap with translations in in~\autoref{tab:morph_oov} and~\autoref{tab:token_translation_recall}, as reported in~\citet{ginn-etal-2023-findings}.

\section{Example Outputs}
\label{sec:outputs}
We include example outputs to show the errors discussed in \S\ref{sec:baselines} and \S\ref{sec:byt5}.
\begin{table*}[!htb]
\small
\centering
\resizebox{\textwidth}{!}{
    \begin{tabular}{cll}
    \toprule
    Sample ID & \\
    \midrule
    \multirow{4}{*}{uspa1245\_136} & transcription & jaan \\
    & translation & Esta bien. \\
    & gold & bueno \\
    & output & esta@bien \\
    \midrule
    \multirow{4}{*}{uspa1245\_149} & transcription & kond (ti') laj chaak \\
    & translation & Cuando en su trabajo. \\
    & gold & cuando ??? PREP trabajo \\
    & output & cu\'ando??? PREP trabajo \\
    \midrule
    \multirow{4}{*}{lezg1247\_1} & transcription & \foreignlanguage{russian}{вич- ни хьун- нва- й къвалах- ар я} \\
    & translation & and all this were real stories \\
    & gold & reflxv- FOC - be- PERF- PST- word- PL was \\
    & output & himself- FOC be- PERF- PTP real.story- PL was \\
    \midrule
    \multirow{4}{*}{lezg1247\_22} & transcription & \foreignlanguage{russian}{ва гьада гьикъван гьа терези- ди- н стрелка пара хкаж хьа- нва- тӀа гьам вилик кутуна- нва} \\
    & translation & she put first that person which is more valuable according to the position of scale arrow.\\
    & gold & and then how that.the.same scale- DIR- GEN arrow very up happened- PERF- COND that before put- PERF \\
    & output & that according.to that.the.same value- OBL- GEN position.in very hit be- PERF- COND that.the.same behind put- PERF \\
    \midrule
    \multirow{4}{*}{arap1274\_991} & transcription & neetotohoe \\
    & translation & Take off your pants ! \\
    & gold & take.off.one's.pants \\
    & output & take.off.pants \\
    \midrule
    \multirow{4}{*}{arap1274\_1998} & transcription & cee'iyo \\
    & translation & payday . \\
    & gold & payment \\
    & output & pay.day \\
    \midrule
    \multirow{4}{*}{arap1274\_1667} & transcription & howouunonetiit hiniito'eibetiit biisiinowoot niihooku'oot beh- 'entou- ' \\
    & translation & pity , relationships , learning through observation , watching closely , it's all there . \\
    & gold & pity.mercy relatedness learning.by.observing watching.along all- located.present 0S \\
    & output & pity.mercy relationships learning.through.observation watching.along all- located.present- 0S \\
    \bottomrule
    \end{tabular}
}
    \caption{Selected example outputs to illustrate errors by \textsc{GlossLM\textsubscript{all}} finetuned models.}
\label{tab:example_outputs}
\end{table*}
\begin{table*}[!htb]
\small
\centering
\resizebox{\textwidth}{!}{
    \begin{tabular}{cllc}
    \toprule
    Sample ID & & & MER \\
    \midrule
    \multirow{6}{*}{lezg1247\_71} & transcription & \foreignlanguage{russian}{Акъадарда и пачагь ибуру ламрал , балкӀандал акъадарда , ламрал акъадарда яда , цӀайни ахъайда , им гьи хуьрей агъуз .} \\
    & gold & mount.ENT this king these-ERG donkey.on horse-SPSS mount.ENT donkey.on mount.ENT or fire.and released.ENT he this village.out.of down \\
    \multirow{2}{*}{  } & \multirow{2}{*}{ByT5} & mount.ENT-ENT this king these-ERG donkey-OBL-SPSS horse-OBL-SPSS mount.ENT-ENT donkey-OBL-SPSS mount.ENT-ENT was fire-FOC & \multirow{2}{*}{0.585} \\ 
    & & \quad put.ENT-ENT he this village-INESS-SPSS down.ENT-ENT \\
    & GlossLM & mount.ENT this king these-ERG donkey.on horse-SPSS mount.ENT donkey.on mount.ENT or fire.and mount.ENT he this village.out.of down & 0.068 \\
    \midrule
    \multirow{5}{*}{lezg1247\_49} & transcription & \foreignlanguage{russian}{Хтана балкӀан , « Гьаа », лагьана « гила чавай физ жеда » лагьана , « пачагьдин руш гъиз ».} \\
    & gold & return-AOR horse Yes say-AOR now help-INELAT go-DAT be-ENT say-AOR king-DIR-GEN girl will.bring-DAT \\
    & ByT5 & coming-AOC horse Yes say-AOC now king-DIR-GEN girl be-ENT say-AOC king-DIR-GEN girl be-ENT  & 0.398 \\ 
    & GlossLM & return-AOR horse Yes say-AOR now help-INELAT go-DAT be-ENT say-AOR king-DIR-GEN girl will.bring-DAT  & 0 \\
    \midrule
    \multirow{5}{*}{lezg1247\_27} & transcription & \foreignlanguage{russian}{АтӀуз гана ибурун кьилерни .} \\
    & gold & cutting.IMC-DAT give-AOR these-ERG-GEN chapter-PL-FOC \\
    & ByT5 & then give-AOR these-ERG-GEN head-PL-FOC  & 0.380 \\ 
    & GlossLM & cutting.IMC-DAT give-AOR these-ERG-GEN chapter-PL-FOC  & 0 \\
    \midrule
    \multirow{5}{*}{lezg1247\_0} & transcription & \foreignlanguage{russian}{И гададини гъил вегьена са жуьт къачуда .} \\
    & gold & this boy-DIR-GEN-ERG hand threw-AOR one pair take-INESS \\
    & ByT5 & this boy-DIR-GEN-ERG hand took-AOC one necklace take-ENT  & 0.327 \\ 
    & GlossLM & this boy-DIR-GEN-ERG hand threw-AOR one pair take-INESS  & 0 \\
    \midrule
    \multirow{5}{*}{lezg1247\_42} & transcription & \foreignlanguage{russian}{И рушни пачагь хьана гила башламишда вичин пачагьвализ .} \\
    & gold & this girl-Q king happened-AOR now started.ENT-INESS himself-ERG-GEN reigned.ENT-ERG-DAT \\
    & ByT5 & this girl-FOC king be-AOC now started.ENT-ENT himself-OBL-GEN reign.to-OBL-DAT  & 0.325 \\ 
    & GlossLM & this girl-Q king happened-AOR now started.ENT-INESS himself-ERG-GEN reigned.ENT-ERG-DAT  & 0 \\
    \bottomrule
    \end{tabular}
}
    \caption{Lezgi examples with the highest difference in MER between finetuned ByT5 and GlossLM outputs.}
\label{tab:lez_example_outputs}
\end{table*}
\begin{table*}[!ht]
    \centering
    \small
    \begin{tabular}{ll||ll||ll}
        \textbf{ID} & \textbf{Avg. Value} & \textbf{ID (cont.)} & \textbf{Avg. Value (cont.)} & \textbf{ID (cont.)} & \textbf{Avg. Value (cont.)} \\ 
        \midrule
        GB020 & 0.183 & GB111 & 0.573 & GB305 & 0.364 \\ 
        GB021 & 0.15 & GB113 & 0.6 & GB309 & 0.185 \\ 
        GB022 & 0.199 & GB114 & 0.502 & GB312 & 0.649 \\ 
        GB023 & 0.058 & GB115 & 0.52 & GB313 & 0.161 \\ 
        GB026 & 0.065 & GB116 & 0.007 & GB314 & 0.038 \\ 
        GB027 & 0.351 & GB117 & 0.705 & GB315 & 0.054 \\ 
        GB028 & 0.373 & GB118 & 0.123 & GB316 & 0.031 \\ 
        GB030 & 0.21 & GB119 & 0.196 & GB317 & 0.03 \\ 
        GB031 & 0.067 & GB120 & 0.276 & GB318 & 0.1 \\ 
        GB035 & 0.484 & GB121 & 0.215 & GB319 & 0 \\ 
        GB036 & 0.023 & GB122 & 0.162 & GB320 & 0.002 \\ 
        GB037 & 0.019 & GB124 & 0.081 & GB321 & 0.065 \\ 
        GB038 & 0.036 & GB126 & 0.369 & GB324 & 0.07 \\ 
        GB039 & 0.165 & GB129 & 0.001 & GB326 & 0.487 \\ 
        GB041 & 0.044 & GB131 & 0.329 & GB327 & 0.563 \\ 
        GB042 & 0.097 & GB132 & 0.519 & GB328 & 0.515 \\ 
        GB043 & 0.021 & GB133 & 0.25 & GB333 & 0.721 \\ 
        GB044 & 0.602 & GB134 & 0.704 & GB334 & 0.276 \\ 
        GB047 & 0.701 & GB135 & 0.729 & GB335 & 0.093 \\ 
        GB048 & 0.634 & GB136 & 0.235 & GB336 & 0.002 \\ 
        GB049 & 0.661 & GB137 & 0.206 & GB408 & 0.352 \\ 
        GB051 & 0.18 & GB138 & 0.43 & GB409 & 0.073 \\ 
        GB052 & 0.02 & GB139 & 0.653 & GB410 & 0.532 \\ 
        GB053 & 0.391 & GB140 & 0.304 & GB415 & 0.233 \\ 
        GB054 & 0.02 & GB147 & 0.567 & GB430 & 0.013 \\ 
        GB057 & 0.167 & GB148 & 0.05 & GB431 & 0.331 \\ 
        GB058 & 0.019 & GB149 & 0.299 & GB432 & 0.263 \\ 
        GB059 & 0.134 & GB151 & 0.018 & GB433 & 0.389 \\ 
        GB068 & 0.484 & GB152 & 0.138 & GB519 & 0.168 \\ 
        GB069 & 0.392 & GB155 & 0.578 & GB520 & 0.094 \\ 
        GB070 & 0.269 & GB156 & 0.032 & GB521 & 0.05 \\ 
        GB071 & 0.326 & GB158 & 0.526 & GB024a & 0.664 \\ 
        GB072 & 0.538 & GB159 & 0.176 & GB024b & 0.132 \\ 
        GB073 & 0.296 & GB160 & 0.6 & GB025a & 0.678 \\ 
        GB074 & 0.543 & GB165 & 0 & GB025b & 0.169 \\ 
        GB075 & 0.556 & GB166 & 0.004 & GB065a & 0.644 \\ 
        GB079 & 0.51 & GB167 & 0.036 & GB065b & 0.184 \\ 
        GB080 & 0.69 & GB170 & 0.452 & GB130a & 0.469 \\ 
        GB081 & 0.032 & GB171 & 0.179 & GB130b & 0.332 \\ 
        GB082 & 0.544 & GB172 & 0.086 & GB193a & 0.626 \\ 
        GB083 & 0.647 & GB177 & 0.272 & GB193b & 0.202 \\ 
        GB084 & 0.498 & GB184 & 0.507 & ~ & ~ \\ 
        GB086 & 0.653 & GB185 & 0.292 & ~ & ~ \\ 
        GB089 & 0.523 & GB186 & 0.101 & ~ & ~ \\ 
        GB090 & 0.146 & GB192 & 0.098 & ~ & ~ \\ 
        GB091 & 0.539 & GB196 & 0.054 & ~ & ~ \\ 
        GB092 & 0.106 & GB197 & 0.041 & ~ & ~ \\ 
        GB093 & 0.394 & GB198 & 0.409 & ~ & ~ \\ 
        GB094 & 0.143 & GB257 & 0.519 & ~ & ~ \\ 
        GB095 & 0.054 & GB260 & 0.097 & ~ & ~ \\ 
        GB096 & 0.018 & GB262 & 0.095 & ~ & ~ \\ 
        GB098 & 0.369 & GB263 & 0.181 & ~ & ~ \\ 
        GB099 & 0.113 & GB264 & 0.062 & ~ & ~ \\ 
        GB103 & 0.368 & GB285 & 0.009 & ~ & ~ \\ 
        GB104 & 0.286 & GB286 & 0.154 & ~ & ~ \\ 
        GB105 & 0.517 & GB291 & 0.013 & ~ & ~ \\ 
        GB107 & 0.521 & GB297 & 0.059 & ~ & ~ \\ 
        GB108 & 0.414 & GB298 & 0.083 & ~ & ~ \\ 
        GB109 & 0.107 & GB299 & 0.314 & ~ & ~ \\ 
        GB110 & 0.163 & GB302 & 0.131 & ~ & ~ \\ 
    \end{tabular}
    \caption{Grambank Feature Averages over Training Set}
    \label{tab:avg_gram}
\end{table*}

\end{document}